\documentclass[conference]{IEEEtran}
\IEEEoverridecommandlockouts
\usepackage{cite}
\usepackage{amsmath,amssymb,amsfonts}
\usepackage{algorithmic}
\usepackage{graphicx}
\usepackage{textcomp}
\usepackage{xcolor}
\usepackage{caption}
\usepackage{booktabs}
\usepackage{url}
\def\BibTeX{{\rm B\kern-.05em{\sc i\kern-.025em b}\kern-.08em
    T\kern-.1667em\lower.7ex\hbox{E}\kern-.125emX}}
    
\begin{document}

\title{Thinking Like Van Gogh: 
Structure-Aware Style Transfer via Flow-Guided 3D Gaussian Splatting
\Large\textit{Seeking ``Exaggeration in the Essential" 
through Geometric Abstraction}}

% \author{
% Anonymous ICME Submission
% }
\author{
\IEEEauthorblockN{Zhendong Wang\IEEEauthorrefmark{1}$^{\triangle}$, 
Lebin Zhou\IEEEauthorrefmark{1}$^{\triangle}$, 
Jingchuan Xiao\IEEEauthorrefmark{2}$^{\triangle}$, 
Rongduo Han\IEEEauthorrefmark{3}, 
Nam Ling\IEEEauthorrefmark{1}, 
Cihan Ruan\IEEEauthorrefmark{1}$^{\blacklozenge}$}
% \IEEEauthorblockA{\IEEEauthorrefmark{0}$^{\triangle}$Equal contribution}
% \IEEEauthorblockA{\IEEEauthorrefmark{0}$^{\blacklozenge}$Corresponding author}
\IEEEauthorblockA{\IEEEauthorrefmark{0} zwang29@scu.edu, josephxky@gmail.com, lzhou@scu.edu, hrd12910@gmail.com, nling@scu.edu, cihanruan@ieee.org}
\IEEEauthorblockA{\IEEEauthorrefmark{1}Department of Computer Science and Engineering, Santa Clara University, Santa Clara, CA, USA}
\IEEEauthorblockA{\IEEEauthorrefmark{2}Department of Mathematics and Computer Studies, Mary Immaculate College, Limerick, Ireland}
\IEEEauthorblockA{\IEEEauthorrefmark{3}College of Software, Nankai University, Tianjin, China}
}

% --- 关键代码开始 ---
\twocolumn[{
  \renewcommand\twocolumn[1][]{#1}
  \maketitle
  \begin{center}
    \centering
    \includegraphics[width=0.9\textwidth]{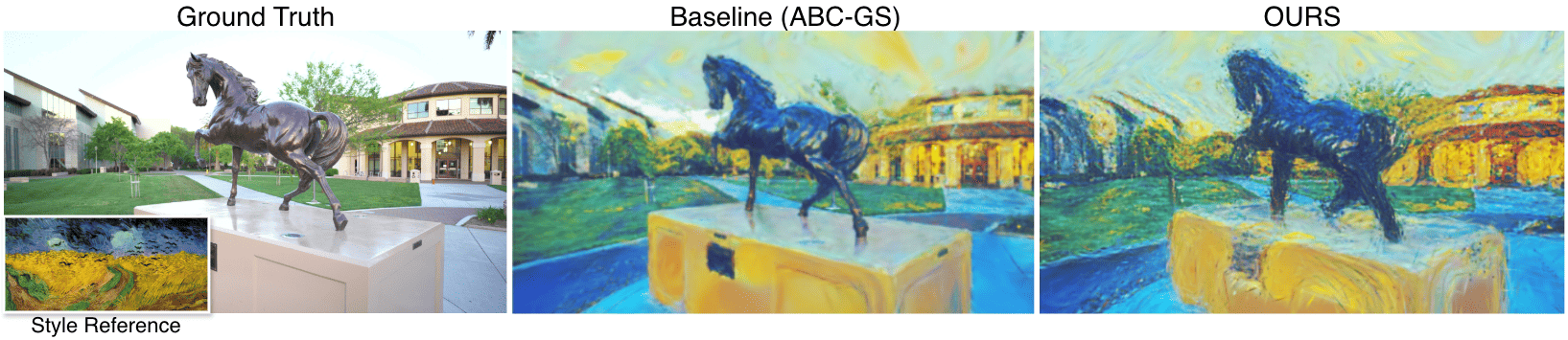} % 替换为你的图片文件名
  \captionof{figure}{\textbf{Subjectivity over Physics.} 
While the baseline method (middle) rigidly preserves photorealistic perspective and lighting—treating style as a flat texture—our method (right) prioritizes subjective geometric flow. We demonstrate that authentic stylization requires sacrificing objective physical fidelity to reconstruct the expressive structural abstraction of the artist.}
    \label{fig:teaser}
  \end{center}
}]

% \maketitle

\renewcommand{\thefootnote}{\fnsymbol{footnote}}
\footnotetext{\textsuperscript{$\triangle$}These authors contributed equally to this work.}
\footnotetext{\textsuperscript{$\blacklozenge$}Corresponding author.}
\renewcommand{\thefootnote}{\arabic{footnote}}

\begin{abstract}
In 1888, Vincent van Gogh wrote, \textit{``I am seeking exaggeration in the essential.''} 
This principle—amplifying structural form while suppressing photographic detail—lies at the core of Post-Impressionist art. 
However, most existing 3D style transfer methods invert this philosophy, treating geometry as a rigid substrate for surface-level texture projection. 
To authentically reproduce Post-Impressionist stylization, geometric abstraction must be embraced as the primary vehicle of expression.

We propose a flow-guided geometric advection framework for 3D Gaussian Splatting (3DGS) that operationalizes this principle in a mesh-free setting. 
Our method extracts directional flow fields from 2D paintings and back-propagates them into 3D space, rectifying Gaussian primitives to form flow-aligned brushstrokes that conform to scene topology without relying on explicit mesh priors. 
This enables expressive structural deformation driven directly by painterly motion rather than photometric constraints.

Our contributions are threefold: 
(1) a projection-based, mesh-free flow guidance mechanism that transfers 2D artistic motion into 3D Gaussian geometry; 
(2) a luminance–structure decoupling strategy that isolates geometric deformation from color optimization, mitigating artifacts during aggressive structural abstraction; 
and (3) a VLM-as-a-Judge evaluation framework that assesses artistic authenticity through aesthetic judgment instead of conventional pixel-level metrics, explicitly addressing the subjective nature of artistic stylization.
\end{abstract}
\noindent\textbf{Code:} \url{https://github.com/zhendong-zdw/TLVG-GS}


\section{Introduction}
\label{sec:intro}
\IEEEPARstart{I}{n} August 1888, Vincent van Gogh wrote from Arles: 
\textit{``I am seeking exaggeration in the essential.''}\cite{vangogh1888} 
This radical declaration—born from his study of Japanese ukiyo-e, 
where economy of line conveyed maximum expression—would catalyze modern 
art's trajectory toward geometric abstraction\cite{silverman2000vangogh}. Van Gogh's Arles period (1888--1890) enacted this 
principle: he abandoned fine academic brushwork for palette knives and 
loaded brushes, creating paint ridges up to 3mm thick\cite{bomford1990} 
that encoded geometry through directional orientation, not photographic 
detail. Wheat fields dissolved into turbulent currents; stars became 
psychological spirals\cite{aragon2006turbulent}. What critics derided as 
``crude'' proved revolutionary—inspiring Post-Impressionism and Expressionism 
to embrace structural syntax over literal representation
\cite{avdeeva2020architectural}.
Each thick, directional stroke functions as a structural vector, 
encoding perceived 3D form through orientation alone (Fig.~\ref{fig:teaser}, 
left, Style Reference).

By systematically eliminating fine detail, Van Gogh compelled viewers to 
perceive volume through directional flow rather than photographic information—what art historians describe as cross-contouring \cite{arnheim1974art}, 
where brushwork aligns with the principal curvature of an object to construct form. 
This insight, which catalyzed Post-Impressionism and Expressionism, 
reveals abstraction not as loss but as transformation: 
a trade from local fidelity toward global geometric coherence 
\cite{yu2025transcendence}. There is a recent computational analysis further confirms that Van Gogh’s rhythmic, directionally coherent brushstrokes 
quantitatively distinguish his works from those of his contemporaries 
\cite{li2021rhythmic}.

% ============================================================
% Para 3: Computational Implication (Streamlined)
% ============================================================
This artistic principle has direct computational implications. 
Recent work demonstrated that parameterized brushstrokes, not pixel-level manipulation, 
capture authenticity in 2D \cite{kotovenko,liu2014tight,bigerelle2023fractal,brachmann2017aesthetics,rajbhandari2024rhythm}—confirming that orientation 
is the syntax of style. Yet extending this to 3D remains unsolved. 
Existing neural style transfer methods, while achieving multi-view consistency 
\cite{abc_gs,fu2019highrelief}, 
fundamentally misunderstand Post-Impressionist painting. 
By treating style as statistical color distributions (Gram matrices)
\cite{gatys2016style},
while rigidly preserving geometric detail, they invert Van Gogh's principle:
they preserve what he eliminated (photographic fidelity) and 
eliminate what he exaggerated (geometric flow)
\cite{sun2022pigments}.

% ============================================================
% Para 4: Our Solution (Compressed)
% ============================================================
In this paper, we propose \textbf{``Thinking Like Van Gogh''}, a flow-guided 
geometric advection framework for 3D Gaussian Splatting (3DGS) that 
realizes Post-Impressionist principles computationally. Our key insight: 
2D directional flow fields in paintings encode the 3D structure artists 
perceived. By extracting these patterns and back-propagating them to 
rectify 3D Gaussians, we transform chaotic point clouds into coherent, 
flow-aligned brushstrokes that wrap around scene topology 
(Fig.~\ref{fig:teaser}, right). This ``exaggerates'' geometric structure 
while ``eliminating'' photographic detail—deliberate abstraction as feature, 
not bug.

% ============================================================
% Para 5: Contributions (One-Line Each)
% ============================================================
To achieve this, we introduce three technical innovations.
First, we develop a mesh-free flow guidance algorithm that ``combs'' 
3DGS primitives into directional brushstrokes via 2D-to-3D back-propagation, 
resolving the ``floating sticker'' artifact without requiring explicit 
topology. Second, we propose luminance-structure decoupling to optimize geometric flow in luminance space while maintaining chromatic consistency, preventing color bleeding during structural deformation.
Third, we pioneer VLM-as-a-Judge evaluation using multiple large multimodal model assessment (ChatGPT, Claude, Gemini, etc.) to assess artistic authenticity beyond pixel-level metrics.

\vspace{-2mm}
\section{Background}
\label{sec:background}
\vspace{-2mm}
\subsection{Related Works}

\begin{figure}[t]
    \centering
    \includegraphics[width=0.9\linewidth]{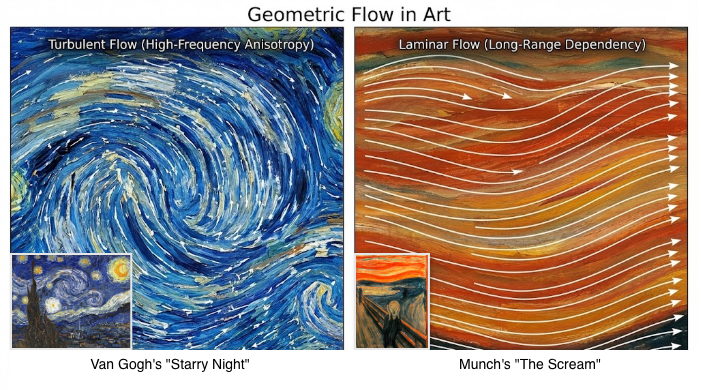} 
    \caption{\textbf{Directional Syntax.}
   Van Gogh (left): turbulent flow. Munch (right): laminar flow. Both prioritize geometric coherence.}
    \label{fig:geometric_flow}
\end{figure}

\begin{figure}[t] 
  \centering
  \includegraphics[width=0.9\linewidth]{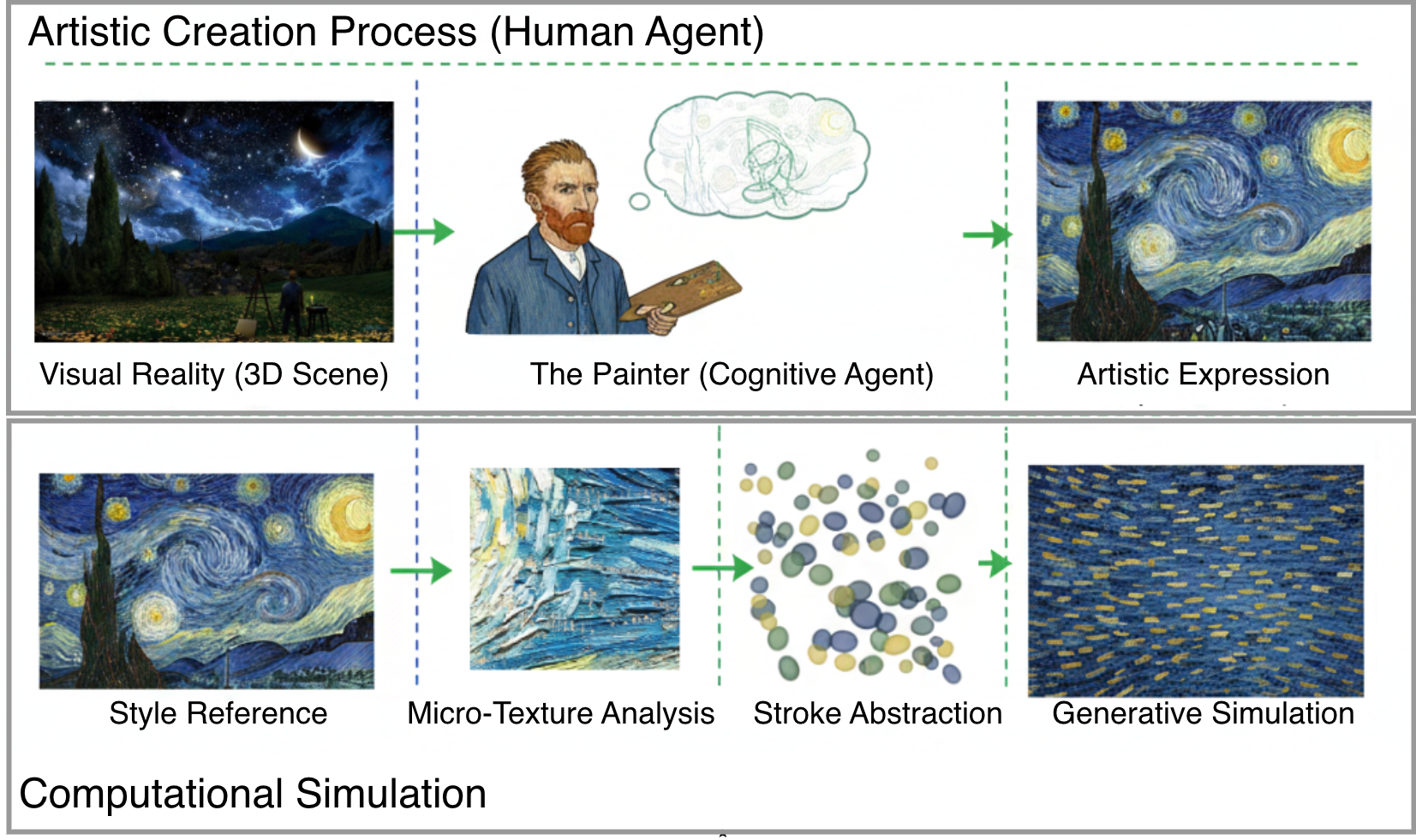}   
  \caption{\textbf{The Analogy between Artistic Cognition and Computational Simulation.}
  Top: Artistic process—perceiving 3D reality, deciding stroke orientations, creating 2D expression. Bottom: Our computational pipeline mirrors this—extracting 
flow from style, modeling as Gaussians, rendering via geometric advection.}
  \label{fig:pipeline_analogy}
\end{figure}

\begin{figure}[t]
    \centering
    \includegraphics[width=0.9\linewidth]{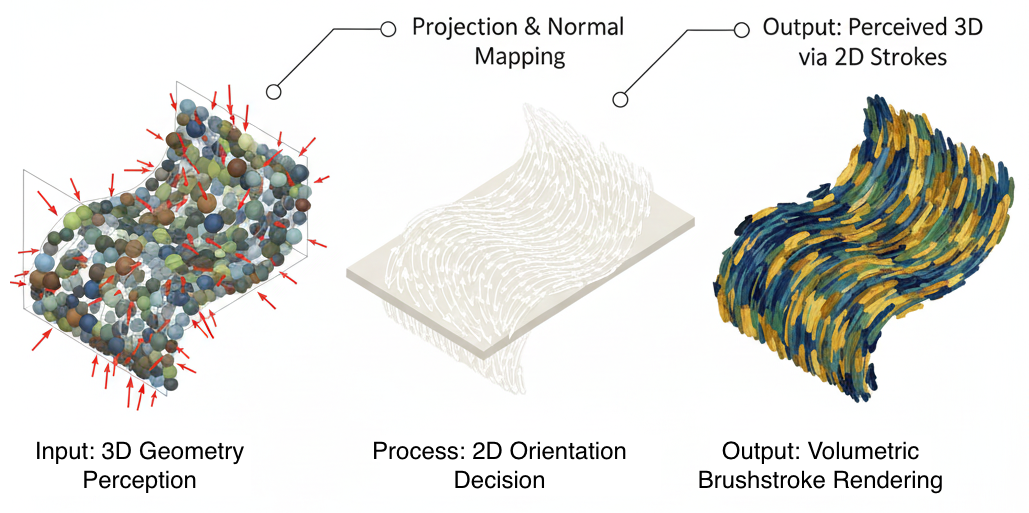} 
    \caption{The Projection-Induced Advection Process.} 
    \label{fig:concept}
\end{figure}

\begin{figure*}[t]
    \centering
    \includegraphics[width=0.75\linewidth]{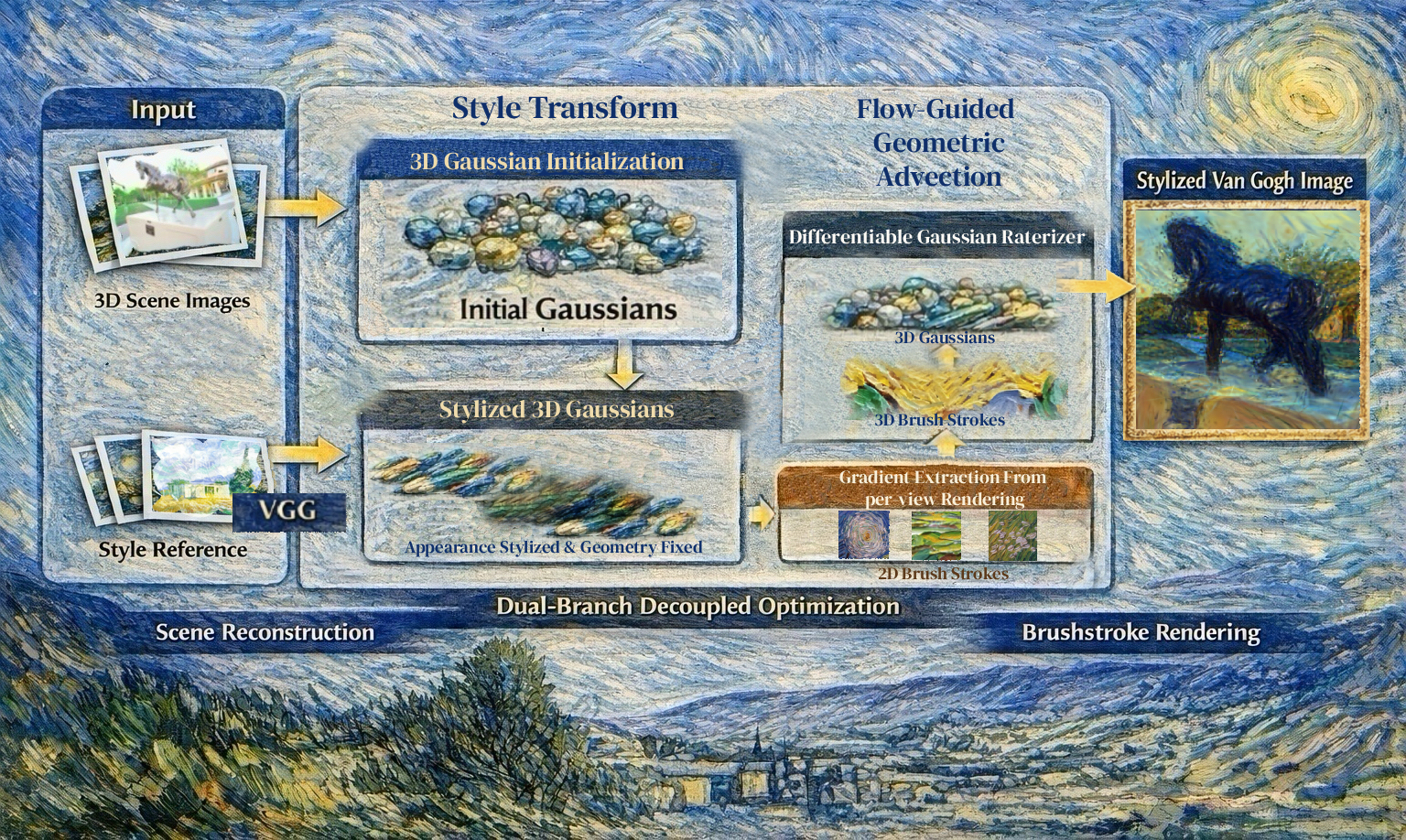} 
    \caption{Overview of the \textbf{Thinking Like Van Gogh} Framework. 
    % Our pipeline introduces a \textbf{dual-branch decoupled optimization} strategy. 
    % The Style Image is pre-processed into orthogonal guidance signals: 2D flow gradients for geometry and luminance channels for structure.
    % Inside the optimization loop, the \textbf{Geometric Advection branch} back-propagates flow gradients to update Gaussian positions ($\mu$) and rotations ($q$), rectifying the brushstroke arrangement.
    % Simultaneously, the \textbf{Luminance-Structure branch} optimizes spherical harmonic colors ($c$) using luminance-only style loss, ensuring clean color synthesis without interfering with geometric deformation.
    % This decoupling allows for significant geometric plasticity while maintaining semantic coherence.
    }
    \label{fig:pipeline}
\end{figure*}

\noindent\textbf{Neural Style Transfer.}
Gatys et al. \cite{gatys2016style} pioneered neural style transfer using 
Gram matrix matching of CNN features, treating style as second-order color 
statistics. Extensions to 3D include NeRF-based methods (ARF \cite{arf}, 
StyleRF \cite{liu2023stylerf}). 
With the advent of 3D Gaussian Splatting, StylizedGS \cite{zhang2025stylizedgs} 
pioneered optimization-based transfer for explicit point clouds. Subsequent works 
focused on efficiency (StyleGaussian \cite{liu2024stylegaussian}) or 
feature disentanglement ($GT^2$-GS \cite{liu2025gt}).
However, despite these advances, current 3DGS approaches (including 
ABC-GS \cite{abc_gs}) predominantly treat style as a texture overlay. 
Even methods that separate geometry from appearance (like $GT^2$-GS) preserve 
the underlying photographic topology rather than actively warping it. This 
``re-texturing'' paradigm fails to capture the structural essence of 
Post-Impressionist painting.

\noindent\textbf{Geometry-Aware Stylization.}
Several works recognize that artistic style involves geometry. Neural 3D 
Strokes \cite{neural3dstrokes} generates stroke primitives but requires 
explicit mesh topology. Geometry Transfer \cite{geomtransfer} aligns depth 
maps but lacks directional flow guidance. Closest to our work, Kotovenko 
et al. \cite{kotovenko} demonstrated that parameterized brushstrokes 
outperform pixel optimization in 2D. We extend this insight to 3D: 
orientation is the syntax of style, and 3DGS primitives must be 
``combed'' into directional arrangements to achieve authentic Post-
Impressionist aesthetics.
\vspace{-2mm}
\subsection{Artistic Motivation: Painting as Dimensional Translation}

\noindent\textbf{The Painter's Cognitive Process:}
Master painters perform a dimensional translation: perceiving 3D scenes 
and encoding volumetric form onto 2D surfaces. Van Gogh achieved this not 
through photographic detail but through directional syntax—
aligning brushstrokes with principal curvature to construct perceived 
geometry (Fig.~\ref{fig:geometric_flow}). His turbulent, high-anisotropy 
strokes encode local curvature; Munch's laminar waves encode psychological 
resonance \cite{aragon2006turbulent}. Both share a geometric principle: 
orientation, not color, encodes 3D structure.

We model this artistic cognition as a three-stage process 
(Fig.~\ref{fig:pipeline_analogy}):
\begin{enumerate}
    \item 3D Perception: 3DGS captures scene geometry via Gaussian 
          positions and covariances—the ``painter's eye''.
    \item Orientation Decision: Extract 2D directional flow from 
          paintings, representing the artist's stroke choices.
    \item Geometric Advection: Back-propagate flow gradients to 
          rectify 3D Gaussians, ``combing'' the point cloud into coherent 
          brushwork—the ``painter's hand''.
\end{enumerate}

\noindent\textbf{The Translation Problem.}
Unlike mesh-based NPR \cite{hertzmann1998painterly}, 3DGS lacks explicit 
surface topology—computing continuous tangent fields is ill-posed. We solve 
this via projection-induced advection (Fig.~\ref{fig:concept}): 
using 2D flow as a proxy to infer optimal 3D Gaussian arrangements, 
analogous to how painters validate 3D perception through 2D rendering.

\vspace{-2mm}
\section{Methodology}
\label{sec:mov}

\subsection{Framework Overview}
\label{sec:overview}
\vspace{-2mm}
Our implementation builds upon ABC-GS \cite{abc_gs}, but diverges in its 
optimization objective. Rather than prioritizing semantic fidelity and 
photorealistic consistency, we focus on \textit{geometric stylization} and 
\textit{painterly abstraction}, aiming to reconstruct the physical logic of 
brushwork.

We formulate stylization as a projection-driven optimization process, in 
which the agent perceives the 3D scene $\mathcal{S}$—represented by a set of 
unstructured 3D Gaussians $\mathcal{G}=\{(\mu_i,q_i,s_i,c_i,\alpha_i)\}$—solely 
through its 2D rendered projections. The optimization jointly minimizes three 
types of energies:
(i) a \textit{flow-alignment energy} that enforces coherent, anisotropic 
brushstroke geometry,
(ii) a \textit{geometric regularization energy} that preserves global 3D 
structure while allowing controlled deformation, and
(iii) an \textit{appearance decoupling energy} that separates structural 
stylization from chromatic consistency.

As illustrated in Fig.~\ref{fig:pipeline}, we realize this objective through 
a dual-branch decoupled optimization strategy:
\begin{itemize}
    \item Geometric Advection Branch: Acting as the painter's ``hand'', this branch utilizes projection analysis to back-propagate 2D flow gradients to the 3D space. It explicitly updates the positions ($\mu$) and rotations ($q$) of the primitives, rectifying the isotropic point cloud into anisotropic constitutive brushstrokes.
    \item Luminance-Structure Branch: Acting as the painter's ``palette'', this branch employs a luminance-only constraint to optimize the color ($c$). This ensures that the geometric deformations do not introduce chromatic artifacts, preserving the purity of the style's palette.
\end{itemize}
This decoupled design enables us to unlock the geometric plasticity of 3DGS necessary for impasto effects, without compromising the semantic coherence of the scene.

\vspace{-2mm}

\subsection{Flow-Guided Geometric Advection}
\label{sec:advection}
Our framework reconstructs the volumetric brushstrokes of master artists (e.g., Van Gogh) through a Geometry-First, Color-Second strategy. Since 3D Gaussian Splatting (3DGS) lacks continuous surface topology (mesh), we cannot rely on pre-computed curvature fields. Instead, we propose a mesh-free approach that rectifies the 3D primitives directly from 2D artistic cues.

\subsubsection{Flow-Aware Primitive Rectification}

Standard 3DGS initializes Gaussians as isotropic primitives, which fails to 
capture the strong directionality of impasto-style brushwork. We therefore 
introduce a flow-alignment energy that rectifies Gaussian orientation 
directly from 2D artistic cues.

We extract a dominant local stroke orientation $v_{\text{2D}}$ from the style reference using structure tensor analysis, where the leading eigenvector of a local gradient covariance matrix is used as a computationally efficient directional proxy in projection space. For each primitive $\mathcal{G}_i$, we define an alignment energy that encourages the projected major axis of the Gaussian to follow the artistic flow:

\begin{equation}
\mathcal{L}_{\text{align}}^{(i)} =
1 - \left|
\left\langle
\Pi\!\left(\mathbf{R}(q_i)\mathbf{e}_1\right),
\mathbf{v}_{\text{2D}}
\right\rangle
\right|
\end{equation}
where $\mathbf{e}_1$ is the canonical major axis of the Gaussian and $\Pi$ 
denotes perspective projection. This formulation is invariant to sign 
ambiguity and measures directional consistency in the image plane. Gaussian rotations $q_i$ are updated by back-propagating the 
gradient of $\mathcal{L}_{\text{align}}$ through the differentiable renderer.

\subsubsection{Gradient-Driven Advection Optimization}

With Gaussian orientations rectified to align with image-space stroke directions,
we further advect the primitives to form coherent volumetric brushstrokes.
Specifically, we unlock the positional parameters $\mu_i \in \mathbb{R}^3$
and update them via gradients of the stylization objective.

Although the stylization losses are defined in the image plane,
their influence on $\mu_i$ is realized through the differentiable projection
operator $\Pi(\cdot)$ that maps 3D Gaussian means to screen-space coordinates.
Let $\mathbf{u}_i = \Pi(\mu_i)$ denote the projected 2D mean.
By the chain rule, the gradient with respect to the 3D position is given by
\begin{equation}
\nabla_{\mu_i} \mathcal{L}
=
\left(
\frac{\partial \Pi(\mu_i)}{\partial \mu_i}
\right)^{\!\top}
\nabla_{\mathbf{u}_i} \mathcal{L}
\end{equation}
where $\nabla_{\mathbf{u}_i} \mathcal{L}$ is the image-space gradient
accumulated by the differentiable renderer.
This Jacobian-mediated backpropagation lifts 2D stylization cues
into consistent 3D updates under perspective projection.

Because all views share the same set of 3D parameters $\{\mu_i\}$,
gradients computed from different viewpoints are accumulated on the same
primitives during optimization.
As a result, the induced advection operates in 3D space rather than
as a view-dependent texture deformation, yielding geometrically coherent
updates across views when multiple viewpoints are considered.

Direct application of these gradients, however, may introduce spurious
motion along the footprint normal of a primitive, causing it to drift
off the underlying surface.
To constrain the optimization to physically plausible advection,
we apply a soft tangential constraint to the optimizer update
$\Delta \mu_i$.
Let $\mathbf{n}_i = \mathbf{R}(q_i)\mathbf{e}_3$ denote the minor-axis direction
of the anisotropic Gaussian, corresponding to the local normal of its footprint.
We project the update onto the associated tangent plane as
\begin{equation}
\Delta \mu_i^{\text{tan}} =
\Delta \mu_i -
\lambda
(\Delta \mu_i \cdot \mathbf{n}_i)\mathbf{n}_i,
\end{equation}
where $\lambda$ controls the strength of the constraint.
This operation suppresses normal-direction drift while preserving motion
along tangential directions, encouraging primitives to slide coherently
along an implicit surface.

In addition, we augment the stylization objective with auxiliary anisotropy
terms defined in the image plane.
These terms penalize footprint energy orthogonal to the stroke direction
while promoting elongation along the stroke tangent.
Through the same Jacobian chain, these image-space anisotropy cues further
bias the 3D advection toward thin, directionally extended volumetric elements
that behave as constitutive brushstrokes across views.

% 4.3 Luminance-Structure Decoupling (颜色保护):

% 问题: 几何变形会导致纹理错位，RGB Loss 会产生脏色块。

\subsection{Luminance-Structure Decoupling for Stable Advection}
\label{sec:color_preservation}

A major challenge in geometric advection is the conflict between structural 
deformation and appearance preservation. When primitives undergo displacement 
to align with brushstroke flow, pixel-level correspondence is disrupted. 
For highly expressive styles like \textit{The Starry Night}, VGG networks 
often misinterpret geometric shifts as texture errors, hallucinating chromatic 
noise that produces a ``muddy'' appearance.

We address this through luminance-structure decoupling (adapted from ARF-Plus \cite{li2025arf}). We hypothesize that 
brushstroke geometry is primarily encoded in luminance, while chromatic 
information should remain stable. We transform images into YIQ space and 
restrict the style loss to the luminance (Y) channel:
\begin{equation}
\mathcal{L}_{\text{style}}
=
\sum_{\ell}
\left\|
\phi_{\ell}(\mathcal{I}_{\text{render}}^{\text{Y}})
-
\phi_{\ell}(\mathcal{I}_{\text{style}}^{\text{Y}})
\right\|_2^2
\end{equation}
where $\phi_{\ell}(\cdot)$ denotes VGG features at layer $\ell$.

To stabilize color appearance, we constrain chromatic statistics in Lab space, 
matching both means $\boldsymbol{\mu}_{ab} = (\mu_a, \mu_b)$ and standard 
deviations $\boldsymbol{\sigma}_{ab} = (\sigma_a, \sigma_b)$:
\begin{equation}
\mathcal{L}_{ab}
=
\frac{1}{2} \left( 
\|\boldsymbol{\mu}_{ab}^{\text{render}} - \boldsymbol{\mu}_{ab}^{\text{ref}}\|_1
+
\|\boldsymbol{\sigma}_{ab}^{\text{render}} - \boldsymbol{\sigma}_{ab}^{\text{ref}}\|_1
\right)
\end{equation}

This decoupling liberates geometric advection from color penalties, enabling 
coherent brushstroke formation without chromatic artifacts.
% ==========================================
% Section 4: Experiments
% ==========================================
\section{Experiments}
\label{sec:exp}

\begin{figure*}[t]
    \centering
    \includegraphics[width=0.98\linewidth]{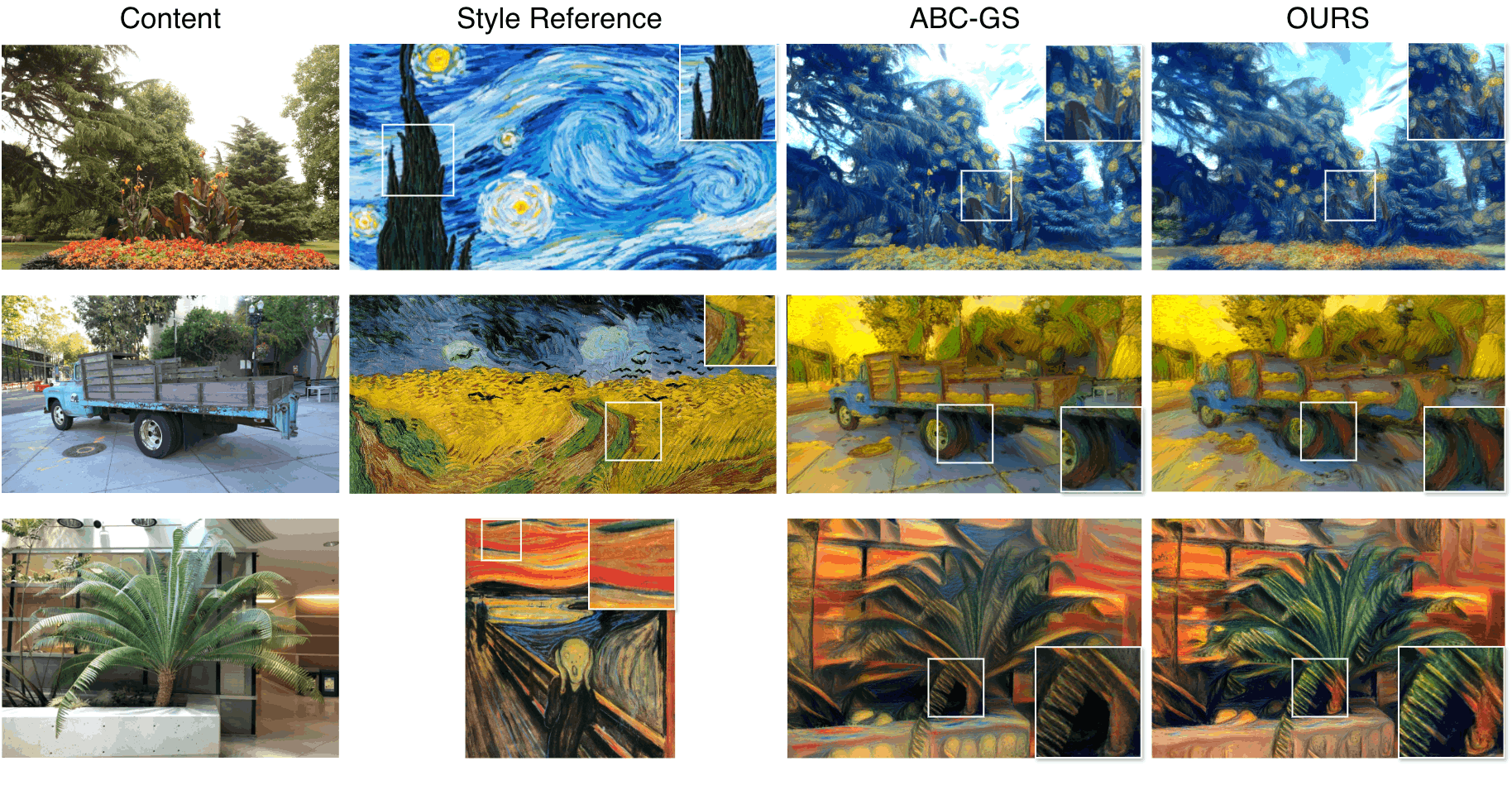}
    \caption{\textbf{Qualitative comparisons with the baseline methods (ABC-GS) using references from Van Gogh and Edvard Munch.} By prioritizing aesthetic energy over physical accuracy, our method captures the creative intent (the ``mind'') of the original masterpiece.}
    \label{fig:visual_result}
\end{figure*}

\subsection{Experimental Setup}
We evaluate our method on real-world scenes (LLFF, Tanks \& Temples) stylized with Post-Impressionist masterpieces (e.g., \textit{The Starry Night}). We benchmark against the state-of-the-art ABC-GS \cite{abc_gs}.
To ensure fair comparison, we adopt the baseline's core configuration, including the VGG-16 backbone \cite{simonyan2014very} and loss weighting scheme.
\textit{Crucially}, unlike ABC-GS which disables density control, we enable \textbf{adaptive densification} (clone and split) to populate geometric voids created by advection.
Experiments were conducted on a single NVIDIA A100 GPU for 3,000 iterations ($\sim$5 mins). Detailed settings are provided in the \textit{Supplementary Material}.

\subsection{Qualitative Evaluation}
We conduct a visual comparison in Fig.~\ref{fig:visual_result}. As shown in the zooms, baseline method ABC-GS effectively transfer color statistics but fail to reconstruct the physical stroke geometry. They treat the object as a smooth surface, resulting in a ``texture mapping'' look where brushstrokes are flatly projected, often cutting across structural edges (e.g., the straight textures on the curved truck wheel).

In contrast, our method produces constitutive brushstrokes. Driven by the projection-induced flow, the Gaussian primitives physically rotate and align with the scene's curvature. This results in a coherent painterly flow that mimics the artist's hand, creating a strong sense of relief and directional energy absent in prior works.

\vspace{-2mm}
\subsection{Quantitative Evaluation}
\label{sec:quantitative}

Quantifying artistic stylization is challenging, as standard metrics often penalize geometric abstraction. We propose a comprehensive protocol combining VLM-based semantic assessment and human user studies.

\subsubsection{The Misalignment of Standard Metrics}
As reported in the \textit{Supplementary Material}, ArtFID scores exhibit high variance contingent on specific style-scene pairs, resulting in comparable averages despite significant perceptual differences. We attribute this instability to the metric's sensitivity to low-level texture statistics rather than geometric coherence. This lack of discriminative consistency necessitates the semantic-aware evaluation introduced below.

\subsubsection{VLM-as-a-Judge: The AI Critic Panel}
\label{sec:vlm_judge}

To overcome the limitations of pixel-based metrics, we established a ``Panel of AI Critics" comprising GPT-5.1, GPT-4o, Claude 4.5, Claude 3.5, Gork and Qwen 3.
We tasked these models to perform randomized pairwise comparisons using a strict ``Tie-Breaker" protocol that explicitly penalizes ``flat sticker artifacts" (texture mapping) while rewarding ``geometric flow." We also recorded an Authenticity Score (1-10) to quantify the gap between digital and physical aesthetics. As detailed in Table \ref{tab:vlm_scores}, the panel reached a strong consensus. Our method achieved an average Win Rate of $>87\%$ across geometric metrics. 

\begin{table}[t]
\centering
\caption{VLM-as-a-Judge Evaluation Result}
\label{tab:vlm_scores}
\small  % 用small字体而不是resizebox
\begin{tabular}{@{}lccc@{}}
\toprule
\textbf{Criterion} & \textbf{Win Rate} & \textbf{Baseline} & \textbf{Ours} \\
\midrule
Flow Alignment & 85.00\% & 7.13 & \textbf{8.38} \\
Materiality    & 85.83\% & 7.13 & \textbf{8.20} \\
Aesthetics     & 85.83\% & 7.31 & \textbf{8.50} \\
\midrule
Average        & 85.83\% & 7.19 & \textbf{8.36} \\
\bottomrule
\end{tabular}
\end{table}

\subsubsection{User Study}
\begin{figure}[t] 
  \centering
  \includegraphics[width=0.95\linewidth]{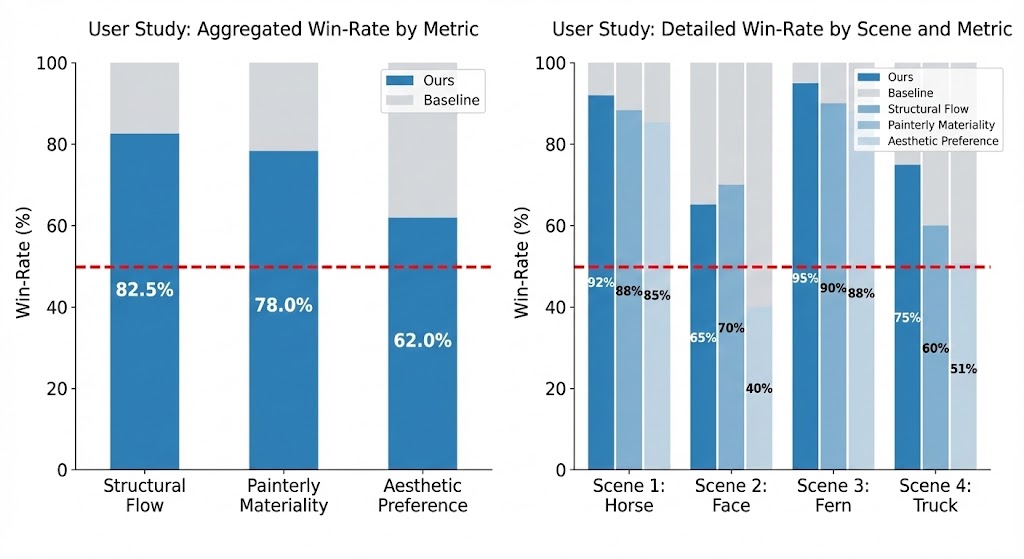}   
  \caption{The Result of User Study}
  \label{fig:user_study}
\end{figure}
To evaluate the perceptual quality of our stylization, we conducted a user study with 30 participants, comprising 18 art experts (professionals in design or fine arts) and 12 general laypeople. We randomly selected 4 diverse groups of scenes for blind pairwise comparisons against baseline methods. For each pair, participants were asked to indicate their preference based on three distinct dimensions: Flow Alignment (geometric logic of brushstrokes), Painterly Materiality (impasto feel and texture), and overall Aesthetic Preference.The aggregated results are presented in Fig.~\ref{fig:user_study}. Compared to baselines, respondents overwhelmingly preferred our method, particularly in geometric metrics, yielding an average win rate of 82\% for Structural Flow and 78\% for Materiality across all scenes. Please refer to the Supplementary Material for detailed methodologies and statistical breakdowns.

\vspace{-2mm}
\subsection{Ablation Study}
\label{sec:ablation}

\begin{figure}[t] 
  \centering
  \includegraphics[width=0.95\linewidth]{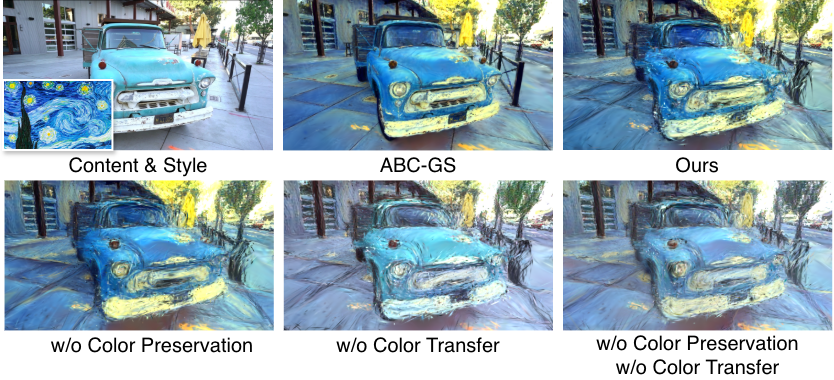}   
  \caption{The Results of Ablation Study}
  \label{fig:ablation}
\end{figure}
\vspace{-2mm}

To validate our framework, we systematically disable key components (Fig.~\ref{fig:ablation}). 
First, disabling Geometric Advection results in flat texture projections lacking volumetric relief, confirming flow guidance is essential for impasto effects. 
Second, replacing Color Preservation with standard RGB loss couples geometry and appearance, causing ``muddy'' chromatic artifacts. 
Finally, enforcing strict regularization (w/o Adaptive Densification) restricts flow magnitude and leads to geometric tearing, whereas our relaxed strategy ensures continuous large-scale deformation. 
Detailed numerical analysis is provided in the Supplement.

\vspace{-2mm}
% ==========================================
% Section 4: Discussion
% ==========================================
\section{Conclusion}
\label{sec:con}

We presented ``Thinking Like Van Gogh'', a flow-guided geometric advection framework for 3DGS that prioritizes directional syntax over texture projection. 
By actively warping geometry to follow artistic flow, we achieve the structural abstraction characteristic of Post-Impressionism—deliberately sacrificing photographic fidelity for expressive coherence.
While we focused on Van Gogh due to his seminal role in defining geometric flow, his influence permeates modern art. 
Future work will extend this ``geometry-first'' paradigm to broader artistic movements (e.g., Expressionism, Futurism), further demonstrating that teaching 3D primitives to follow the syntax of orientation is key to the computational replication of artistic cognition. Additional experimental results, theoretical analysis, and implementation details
are provided in the supplementary material.

\vspace{-2mm}
\bibliographystyle{IEEEtran}

\bibliography{Refs}

\end{document}